\documentclass[10pt,twocolumn,letterpaper]{article}

\usepackage{cvpr}              % To produce the CAMERA-READY version
\usepackage[utf8]{inputenc} % allow utf-8 input
\usepackage[T1]{fontenc}    % use 8-bit T1 fonts
\usepackage{url}            % simple URL typesetting
\usepackage{booktabs}       % professional-quality tables
\usepackage{amsfonts}       % blackboard math symbols
\usepackage{nicefrac}       % compact symbols for 1/2, etc.
\usepackage{microtype}      % microtypography
\usepackage[table]{xcolor} % gives \rowcolors, \rowcolor, and color mixing like gray!10
\usepackage{graphicx}
\usepackage{amsmath}
\usepackage{float}
\usepackage{booktabs}
\usepackage{longtable}
\usepackage{array}
\usepackage{placeins}

\newcolumntype{L}[1]{>{\raggedright\arraybackslash}p{#1}}
\usepackage{tabularx} % Add this in your preamble
\usepackage[most]{tcolorbox}
\usepackage{tabularx}
\usepackage{booktabs}
\usepackage{array}
\usepackage{longtable}

\usepackage{booktabs}
\usepackage{array}
\usepackage{makecell}
\usepackage{longtable}
\usepackage{caption}
\usepackage{etoolbox}

\newcolumntype{L}[1]{>{\raggedright\arraybackslash}p{#1}}

\definecolor{cvprblue}{rgb}{0.21,0.49,0.74}
\usepackage[pagebackref,breaklinks,colorlinks,allcolors=cvprblue]{hyperref}

\def\paperID{AIGENS-10} % *** Enter the Paper ID here
\def\confName{CVPR}
\def\confYear{2026}

\title{On the Robustness of Diffusion-Based Image
Compression to Bit-Flip Errors}

\usepackage{xcolor}

\author{
Amit Vaisman\thanks{Equal contribution}\\
Department of Computer Science\\
Technion - Israel Institute of Technology\\
{\tt\small amit.vaisman@campus.technion.ac.il}
\and
Gal Pomerants\footnotemark[1]\\
Department of Computer Science\\
Technion - Israel Institute of Technology\\
{\tt\small galkesten@campus.technion.ac.il}
\and
Raz Lapid\\
Deepkeep\\
Tel Aviv, Israel\\
{\tt\small raz.lapid@deepkeep.ai}
}

\usepackage{amsmath,amsfonts,bm,upgreek}

\def\eqref#1{eq.~(\ref{#1})}
\def\1{\bm{1}}

\def\rvx{{\mathbf{x}}}

\def\rvz{{\mathbf{z}}}

\def\vmu{{\bm{\mu}}}

\def\vs{{\bm{s}}}

\def\vz{{\bm{z}}}

\def\mC{{\bm{C}}}

\def\mI{{\bm{I}}}

\DeclareMathAlphabet{\mathsfit}{\encodingdefault}{\sfdefault}{m}{sl}
\SetMathAlphabet{\mathsfit}{bold}{\encodingdefault}{\sfdefault}{bx}{n}

\DeclareMathOperator*{\argmax}{arg\,max}
\DeclareMathOperator*{\argmin}{arg\,min}

\usepackage[capitalize,noabbrev]{cleveref}

\begin{document}

\maketitle

\begin{abstract}

Modern image compression methods are typically optimized for the rate--distortion--perception trade-off, whereas their robustness to bit-level corruption is rarely examined. We show that diffusion-based compressors built on the Reverse Channel Coding (RCC) paradigm are substantially more robust to bit flips than classical and learned codecs. We further introduce a more robust variant of Turbo-DDCM that significantly improves robustness while only minimally affecting the rate--distortion--perception trade-off. Our findings suggest that RCC-based compression can yield more resilient compressed representations, potentially reducing reliance on error-correcting codes in highly noisy environments.  A reference implementation is available at \href{https://github.com/galkesten/Diffusion-Compression-Editing-Attacks}{\includegraphics[height=1em]{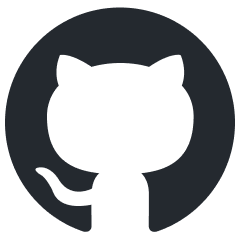}}.
\end{abstract}    
\section{Introduction}
\label{sec:intro}

% intro to image compression
The field of image compression has undergone a significant transformation in recent years, shifting toward neural-based approaches \cite{balle2017end,balle2018variational,minnen2018joint,he2022elic}. These methods have enabled a new regime of increased compression, reaching very low bitrates compared to the non-neural methods, while maintaining strong perceptual quality \cite{agustsson2019generative,mentzer2020high}. More recently, diffusion models have emerged as a powerful paradigm for image compression, leveraging learned priors over the natural image manifold \cite{yang2023lossy,hoogeboom2023high,careil2023towards}. They have been employed through dedicated end-to-end training \cite{yang2023lossy}, adaptation or reuse of pre-trained/foundation generative models \cite{careil2023towards,relic2024lossy}, and even in zero-shot settings \cite{elata2024psc, vaisman2025turboddcmfastflexiblezeroshot, ohayon2025compressed, vonderfecht2025lossy}. Collectively, these approaches represent some of the strongest current results on the rate--distortion--perception trade-off in perceptual image compression \cite{pmlr-v97-blau19a,hoogeboom2023high,relic2024lossy}.

% bitflips
However, real-world systems face practical challenges that prevent the full utilization of compression methods. One such challenge is the occurrence of bit-flip errors. Bit-flips may affect compressed representations in several scenarios. First, errors can arise during transmission over communication networks, for example due to noisy channels \cite{bourtsoulatze2019deep,naseri2024deep}. Second, when compressed data is stored, bit-flips may occur as a result of hardware degradation or memory faults over time \cite{cai2015data}. Finally, bit-flips may also be induced intentionally through adversarial attacks. For instance, attackers can manipulate stored bit representations in memory using a row-hammer attack \cite{row-hammer, rakin2019bit}, or deliberately perturb the communication channel to disrupt data transmission \cite{kim2021channel}.

\begin{figure*}[t]
    \centering
    \includegraphics[width=1\textwidth]{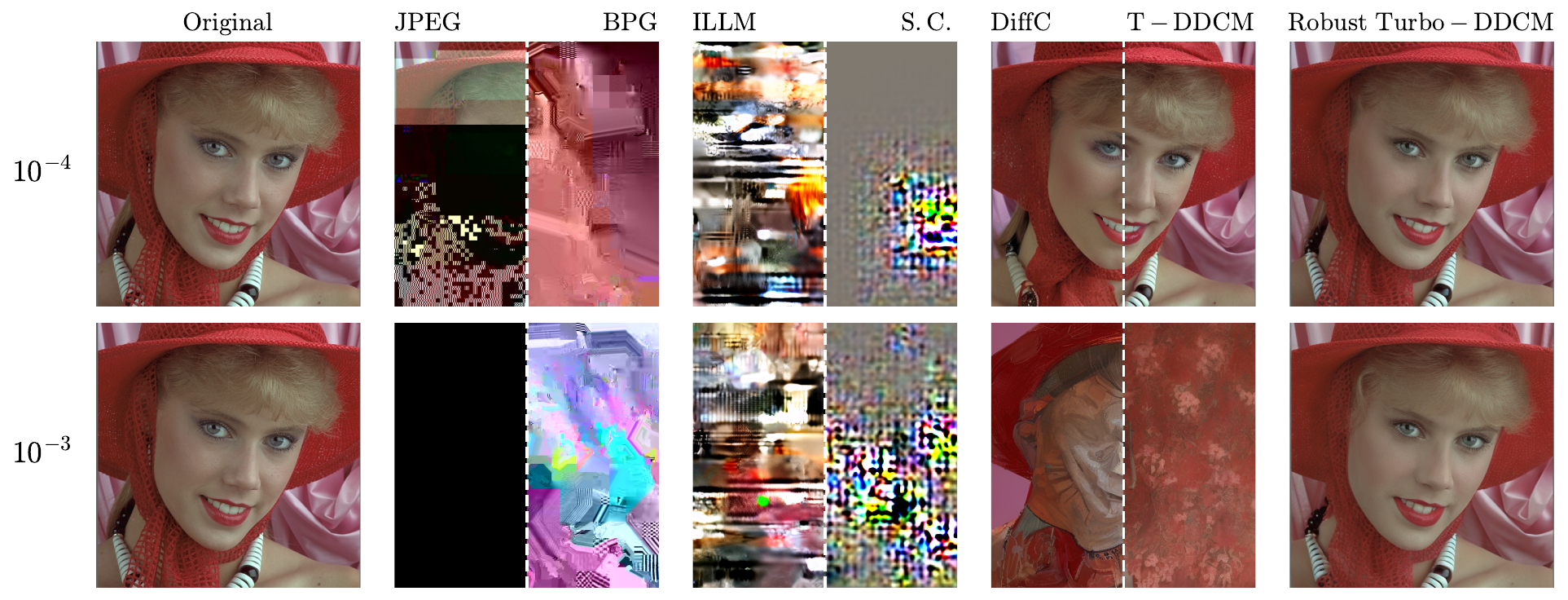}
    \caption{\textbf{Image Compression Methods Robustness to Bit-Flip Errors:} The figure shows reconstructed images from the Kodak24 dataset after transmission through a noisy channel with bit-flip probabilities of $10^{-4}$ (first row) and $10^{-3}$ (second row). S.C. denotes StableCodec, and T-DDCM denotes Turbo-DDCM. At a bit-flip probability of $10^{-4}$, RCC-based methods (the last three) perform well, whereas the others degrade significantly. At $10^{-3}$, our proposed method, Robust Turbo-DDCM, is the only approach that maintains good reconstruction quality. All methods operate at bit-per-pixel rate, yielding high-fidelity reconstructions in a noise-free channel.}
    \label{fig:teaser}
\end{figure*}

As illustrated in Fig.~\ref{fig:teaser}, even a small number of bit-flips in the compressed representation may significantly degrade reconstruction quality, and in extreme cases may render the file undecodable. To mitigate such errors, practical systems typically employ error-correcting codes (ECC). However, ECC increases the size of the compressed representation and consequently degrades rate-distortion-perception \cite{choi2019neuraljointsourcechannelcoding}.

In this work, we ask a fundamental question: can diffusion-based image compression deliver increased robustness as well as increased compression? We first show empirically that zero-shot diffusion-based compressors built on the reverse channel coding (RCC) paradigm \cite{theis2022algorithmscommunicationsamples} are substantially more resilient to bit-flip errors than both classical (non-neural) codecs and trained neural compression methods, maintaining perceptual quality under corruption levels that severely degrade competing bitstreams.

We then propose a robust variant of Turbo-DDCM \citep{vaisman2025turboddcmfastflexiblezeroshot}, a recently introduced zero-shot diffusion-based compression method. Our approach, \emph{Robust Turbo-DDCM}, achieves state-of-the-art robustness to bit flips with only a minor impact on the rate–distortion–perception trade-off (see Appendix \ref{app:turbo-ddcm-rate-distortion-tradeoff}). Under highly noisy channel conditions, our method yields near-identical reconstructions, whereas competing approaches, including the vanilla Turbo-DDCM, exhibit pronounced artifacts or fail to preserve comparable quality.

Together, these findings show that diffusion-based compression can offer not only increased compression, but also strong robustness to bit-level corruption. 

\section{Related Work \& Background}
\label{sec:related}

\subsection{Lossy Image Compression}
Lossy image compression aims to minimize the number of bits required to represent an image while allowing irreversible information loss. Classical codecs, such as JPEG \cite{wallace1991jpeg} and BPG \cite{bellard2018bpg}, rely on handcrafted transforms, quantization, and entropy coding. In contrast, many deep learning-based methods learn the compression process directly, using architectures such as variational autoencoders, GANs, and diffusion models \cite{balle2017end,balle2018variational,minnen2018joint,agustsson2019generative,mentzer2020high,yang2024lossy,relic2024lossy}. More recently, pre-trained diffusion models have been employed in a zero-shot manner for image compression. These include PSC \cite{elata2024psc}, which constructs an image-adaptive transform using a pre-trained diffusion model, as well as RCC-based methods such as DiffC \cite{vonderfecht2025lossy}, DDCM \cite{ohayon2025compressed}, and Turbo-DDCM \cite{vaisman2025turboddcmfastflexiblezeroshot}, which transmit information that explicitly steers the denoising process toward a target image.

\subsection{Bit-Flip Errors}
Bit-flip errors (BFEs) occur when a stored or transmitted bit is unintentionally inverted from 0 to 1 or from 1 to 0. Such errors may arise from transmission noise, hardware faults, memory degradation, or external interference \cite{mackay2003information,proakis2008digital}. In information theory, these effects are typically modeled using a \emph{noisy channel}, where transmitted symbols may be corrupted before reaching the receiver \cite{shannon2001mathematical}.

To mitigate such errors, communication pipelines typically integrate compression with error protection. Data is first compressed to reduce redundancy and subsequently protected using ECC, which introduces controlled redundancy that enables the receiver to detect or correct corrupted bits \cite{choi2019neuraljointsourcechannelcoding,naseri2024deep}. This strategy is widely used in practical communication standards such as Wi-Fi \cite{ieee80211}, where poor channel conditions may require stronger protection, sometimes adding roughly one redundant bit for every information bit transmitted.

In the standard pipeline, where compression is followed by separate error correction, many compression algorithms rely on variable-length entropy coding, such as Huffman or arithmetic coding, to achieve low bitrates \cite{mackay2003information,proakis2008digital}. In such schemes, a single bit error can disrupt decoding, cause loss of synchronization, and propagate corruption across many subsequent symbols \cite{Hussain2018ImageCT}. By contrast, diffusion RCC-based compression methods such as DDCM and Turbo-DDCM can achieve very low bitrates without additional entropy coding, raising the question of whether introducing entropy coding provides a worthwhile benefit if these methods are more robust to bit-flip errors.

\subsection{DDCM and Turbo-DDCM}
Denoising diffusion probabilistic models (DDPMs) generate samples through a reverse diffusion process~\citep{ho2020denoising}. At each timestep the model predicts the mean of the reverse transition and adds Gaussian noise,
\begin{equation}
\label{eq:backward_ddpm}
\rvx_{t-1} = \vmu_t(\rvx_t) + \sigma_t \rvz_t,
\qquad
\rvz_t \sim \mathcal{N}(\mathbf{0}, \mI),
\end{equation}
for $t = T+1, \ldots, 2$, while no noise is added at the final step $t=1$.

Denoising diffusion codebook model (DDCM)~\citep{ohayon2025compressed} reformulates this stochastic reverse transition by replacing the Gaussian noise with a structured selection from reproducible codebooks $\mC_t$. Each codebook $\mC_t$ contains $K$ i.i.d.~Gaussian noise vectors (``atoms''),
\begin{equation}
\mathcal{C}_t =
\left[\vz_t^{(1)}, \vz_t^{(2)}, \ldots, \vz_t^{(K)}\right],
\qquad
t = 2, \ldots, T+1,
\end{equation}
and the reverse step becomes
\begin{equation}
\label{eq:backward_ddcm}
\rvx_{t-1} = \vmu_t(\rvx_t) + \sigma_t \mC_t(k_t),
\end{equation}
where $k_t \sim \mathrm{Unif}(\{1,\ldots,K\})$.

A key property of DDCM is that it enables zero-shot image compression. Given a target image $\rvx_0$, the encoder computes the denoising residual between $\rvx_0$ and the MMSE estimate $\hat{\rvx}_{0|t}$ at each timestep $t$, and selects the atom with maximal correlation with this residual,
\begin{equation}
\label{eq:ddcm_objective}
k_t =
\argmax_{k \in \{1,\ldots,K\}}
\langle
\mC_t(k), \rvx_0 - \hat{\rvx}_{0|t}
\rangle .
\end{equation}

The resulting sequence of indices $(k_t)_{t=2}^{T+1}$ forms the compressed representation. During decompression the same reverse process (Eq.~\ref{eq:backward_ddcm}) is executed deterministically using the stored indices. Compression rate is measured in bits per pixel (BPP), defined as the number of transmitted bits divided by the number of image pixels. The resulting bitrate of DDCM is

\begin{equation}
\label{eq:ddcm_bpp}
\mathrm{BPP}_{\mathrm{DDCM}}
=
\frac{T\lceil\log_2 K\rceil}{\text{number of pixels}} .
\end{equation}

Turbo-DDCM~\citep{vaisman2025turboddcmfastflexiblezeroshot} builds on DDCM and introduces a more efficient sparse approximation. At each timestep $t$, the residual is approximated using a sparse linear combination of $M$ codebook atoms obtained by solving
\begin{equation}
\label{eq:turbo_objective}
\begin{aligned}
\vs_t^* = \argmin_{\vs_t \in \mathbb{R}^K}
\left\| \mC_t \vs_t - (\rvx_0 - \hat{\rvx}_{0|t}) \right\|_2^2, \\
\text{s.t. } \|\vs_t\|_0 = M,\ \forall i,\ (\vs_t)_i \in \mathcal{V} \cup \{0\}.
\end{aligned}
\end{equation}
Here $\vs_t$ is a sparse coefficient vector indicating the selected atoms. Due to the near-orthogonality of Gaussian atoms in high-dimensional spaces, this optimization admits an efficient closed-form thresholding solution that avoids the iterative search required by matching pursuit. The resulting noise is
\begin{equation}
\label{eq:turbo_noise}
\vz_t^* =
\frac{\mC_t \vs_t^*}{\mathrm{std}(\mC_t \vs_t^*)},
\end{equation}
which replaces the stochastic noise in Eq.~\ref{eq:backward_ddpm}.

Turbo-DDCM also introduces a more efficient bitstream protocol to further reduce the bitrate. At each timestep, the encoder transmits an \emph{unordered} combination of $M$ atoms from the $K$-atom codebook, together with their quantized coefficients, each encoded using $C$ bits, for a total of $\left\lceil \log_2 \left( \binom{K}{M} \right) \right\rceil + MC$ bits compared to the original. In addition, the final $N$ denoising steps are replaced by deterministic DDIM updates, which act as refinement steps and require no transmitted noise information. Hence, only the first $T-N-1$ steps contribute to the bitrate.
\begin{equation}
\label{eq:turbo_bpp}
\mathrm{BPP}_{\mathrm{Turbo\text{-}DDCM}}
=
\frac{(T-N-1)
\left(
\left\lceil \log_2 \left( \binom{K}{M} \right) \right\rceil + MC
\right)}
{\text{number of pixels}} .
\end{equation}

\section{Bit-flips Robustness Analysis}
\label{sec:bit_flips_robust_anslysis}
RCC-based diffusion compression methods represent an image indirectly, by encoding control signals that guide the denoising trajectory toward the target image rather than storing pixel values or transform coefficients directly. Because the compressed representation affects reconstruction through this iterative generative process, we hypothesize that such methods may exhibit some tolerance to bit errors, so that a limited number of bit flips does not necessarily cause catastrophic reconstruction failure. For example, in Turbo-DDCM, the bitstream specifies, at each denoising step, a sparse approximation of the noise used to steer the reverse diffusion process. Small perturbations in this representation may still yield a similar steering signal and therefore a similar reconstruction trajectory.

To test this hypothesis, we evaluate the robustness of several image compression methods to bit flips by simulating transmission through a Binary Symmetric Channel (BSC) \cite{shannon2001mathematical}. In this model, each bit in the compressed bitstream is independently flipped with probability $p$, known as the bit error rate (BER), i.e., the probability that a transmitted bit is received incorrectly. For each compression method, we corrupt the compressed bitstream according to this model, decode it using the corresponding decoder, and measure the resulting reconstruction quality. We evaluate BER values of $10^{-6}, 10^{-5}, 10^{-4}, 10^{-3}, 10^{-2}$, and $10^{-1}$. For each image and BER value, we repeat the corruption and decoding process 10 times.

To evaluate reconstruction quality, we report metrics capturing both distortion and perceptual quality. Distortion is measured using PSNR and LPIPS~\citep{zhang2018perceptual}, while perceptual quality is evaluated using FID~\citep{bińkowski2018demystifying}.

In some cases, bit flips in the compressed bitstream may produce corrupted files that cannot be decoded by the corresponding decoder. Such cases are excluded from the image-quality metrics, and the fraction of corrupted files is reported as an additional robustness measure (see Sec.~\ref{sec:experiments}).

% \subsection{Robust Turbo-DDCM}
% \label{sec:robust_turbo_ddcm}
% When examining the bit-protocol of Turbo-DDCM, it becomes clear that the portion encoding the lexicographic order of the chosen atoms is highly sensitive to bit flips. A single bit flip can alter many atoms, resulting in a substantially different constructed noise. For example, with $K=8$ and $M=3$, a lexicographic index of 0 corresponds to the codebook selection $\{0,1,2\}$. Flipping the most significant bit changes the index to 32, producing the selection $\{1,4,7\}$. In contrast, a bit flip in the coefficients portion affects only a single atom in the constructed noise. Thus, in Robust Turbo-DDCM we encode each atom separately, as a number between 0 to $K$. As a result, the bits-per-pixel (BPP) is given by:
% \begin{align}
%     \text{BPP}_\text{Robust Turbo-DDCM} = \frac{(T-1-N)(\lceil \operatorname{log}_2(K) \rceil M + MC)}{\text{number of pixels}}.
% \end{align}

% \raz{What is C in the equation above? define.}

% It is important to note that this protocol has a higher BPP compared to the original Turbo-DDCM protocol \todo{APP?}. However, as presented in Section~\ref{sec:experiments}, the new protocol is much more robust to bit-flips.

\section{Robust Turbo-DDCM}
\label{sec:robust_turbo_ddcm}
When examining the bit-protocol of Turbo-DDCM, it becomes apparent that the portion encoding the lexicographic order of the chosen atoms is highly sensitive to bit flips. In the original protocol, the selected subset of atoms is encoded as a single lexicographic index representing one combination out of $\binom{K}{M}$ possibilities. Consequently, a single bit flip in this index may change the decoded combination entirely, resulting in a substantially different set of atoms and therefore a significantly different constructed noise.

For example, with $K=8$ atoms and selecting $M=3$ atoms, the lexicographic index $0$ corresponds to the atom set $\{0,1,2\}$. Flipping the most significant bit changes the index to $32$, which corresponds to the atom set $\{1,4,7\}$. Thus, a single bit flip can alter multiple atoms simultaneously, leading to large reconstruction errors.

In contrast, bit flips in the coefficient portion affect only the corresponding atom coefficient and therefore have a much more localized effect on the constructed noise.

To mitigate this failure mode, we propose \emph{Robust Turbo-DDCM}, which encodes each selected atom independently rather than jointly via a lexicographic index. Specifically, each atom index is encoded separately as an integer in $\{0,\dots,K-1\}$. As a result, a bit flip can only corrupt the index of a single atom rather than the entire subset selection. This significantly improves robustness to bit errors.

The cost of this modification is an increase in the bit-rate, since encoding atoms independently requires $\lceil \log_2(K) \rceil$ bits per atom instead of the more compact lexicographic encoding. Denoting by $C$ the number of bits used to encode each atom coefficient, the BPP of Robust Turbo-DDCM is

\begin{align}
\text{BPP}_{\text{Robust Turbo-DDCM}} =
\frac{(T-1-N)\left(M\lceil \log_2(K) \rceil + MC\right)}{\text{number of pixels}}.
\end{align}
Here, $\lceil \log_2(K) \rceil$ bits encode the index of each selected atom from M atoms and $C$ bits encode its corresponding coefficient.

\begin{figure*}[t]
    \centering
    \includegraphics[width=0.7\textwidth]{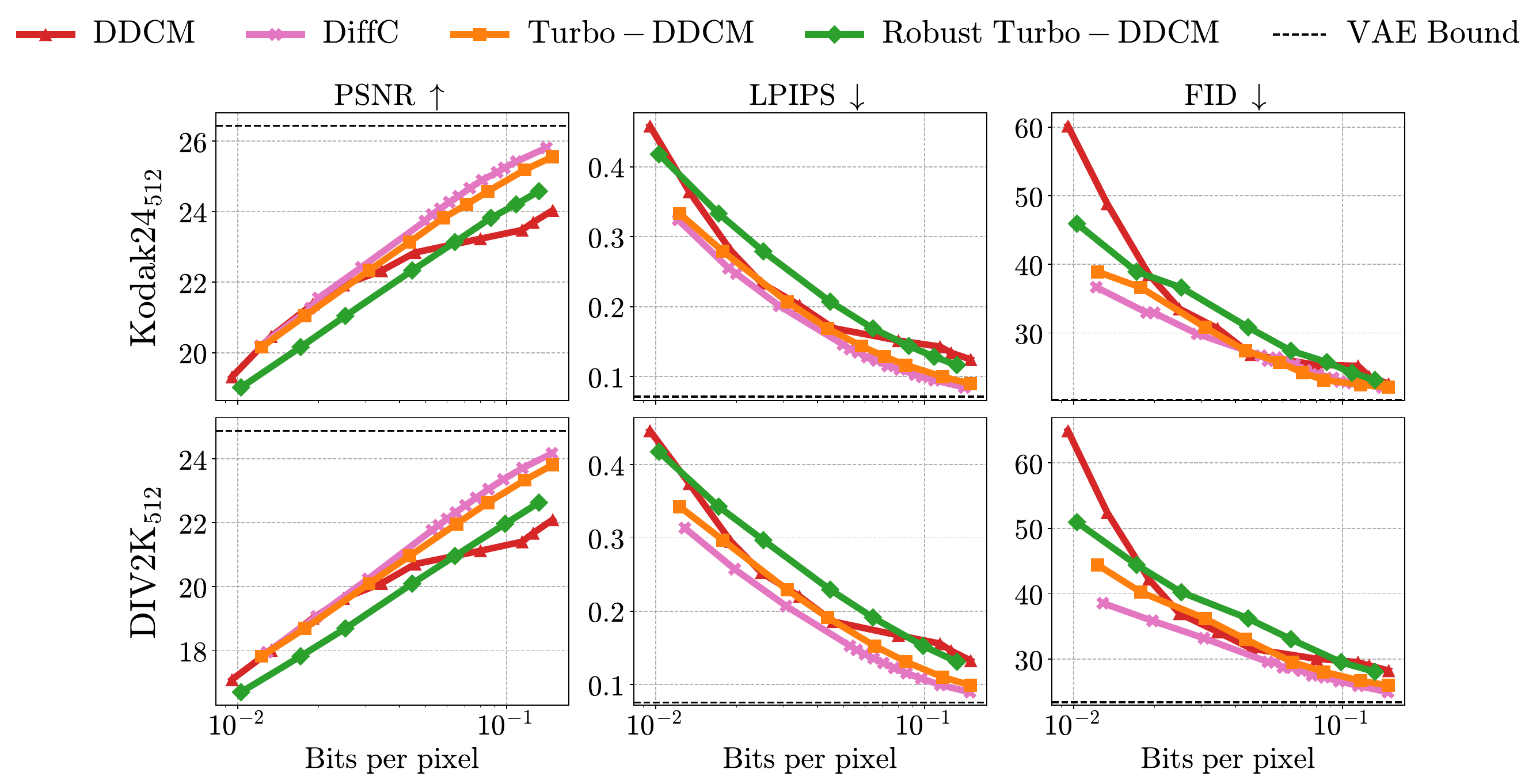}
    \caption{\textbf{Rate-Distortion-Perception Tradeoff:} Robust Turbo-DDCM increases the bits per pixel compared to Turbo-DDCM. While it significantly improves robustness, it exhibits inferior rate–distortion–perception performance.}
    \label{fig:rate_distortion_perception}
\end{figure*}

Although the robust protocol requires more bits per atom than the original Turbo-DDCM encoding, this does not necessarily lead to a large quality gap under a fixed bit budget. Reconstruction quality improves with the number of selected atoms $M$, but with diminishing returns (see Fig. ~\ref{fig:rate_distortion_perception}). Thus, while the original protocol can encode more atoms for the same bit budget, the resulting quality gain is limited, whereas the robust protocol provides substantially improved resilience to bit flips as presented in Section~\ref{sec:experiments}. This creates a direct trade-off between quality and robustness, allowing the choice of protocol to depend on the application requirements.

\section{Experiments}
\label{sec:experiments}

\begin{figure*}[t]
    \centering
    \includegraphics[width=0.9\textwidth]{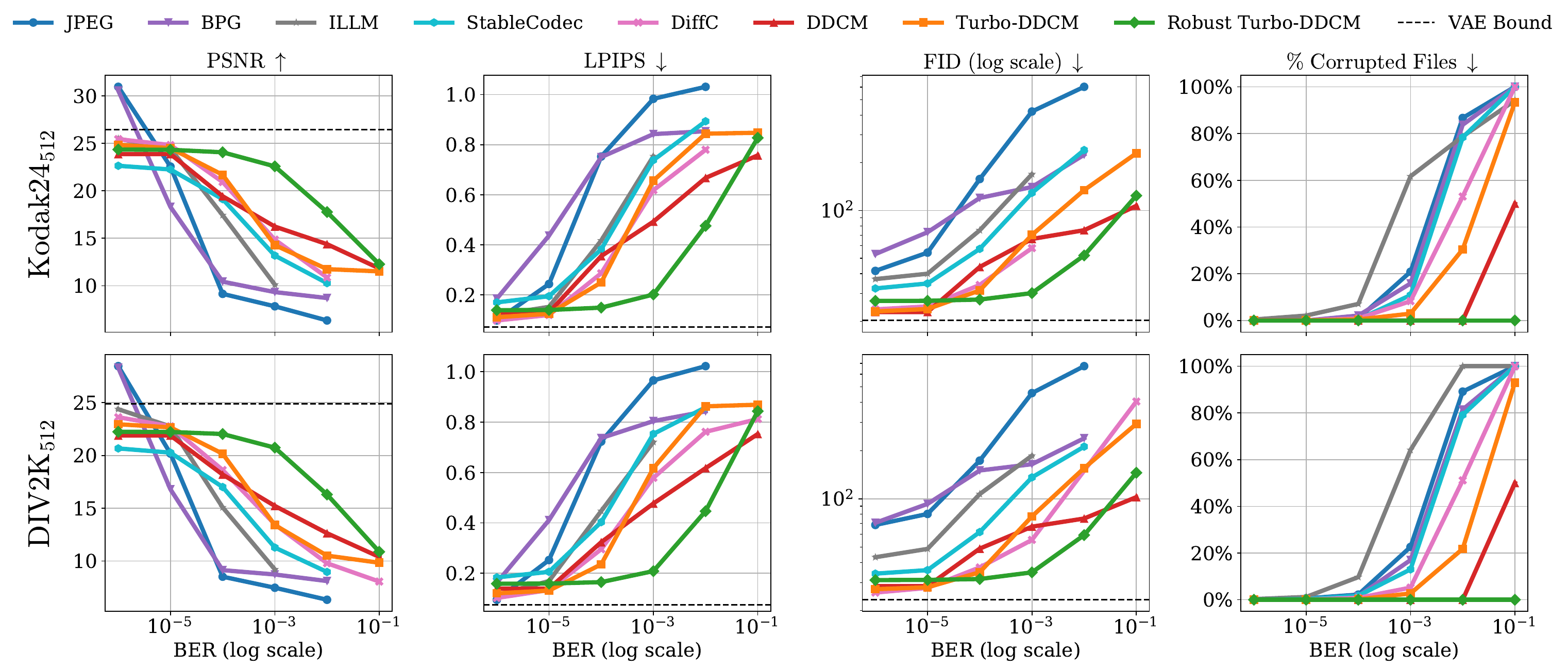}
    \caption{\textbf{Quantitative Results:} We compare the bit-flip robustness of several image compression methods, including non-neural, trained neural, and diffusion-based RCC methods, by evaluating distortion (PSNR or LPIPS) and perception (FID) across multiple BER values. Overall, RCC-based methods outperform the others, with Robust Turbo-DDCM demonstrating particularly strong robustness. The dashed vertical line in each subplot indicates the encoder–decoder distortion bound imposed by SD~2.1, obtained by passing clean images through the encoder–decoder without compression. Since all compared RCC-based methods rely on SD~2.1, they are all subject to this distortion bound.}
    \label{fig:quant_main}
\end{figure*}

\begin{figure*}[t]
    \centering
    \includegraphics[width=0.8\textwidth]{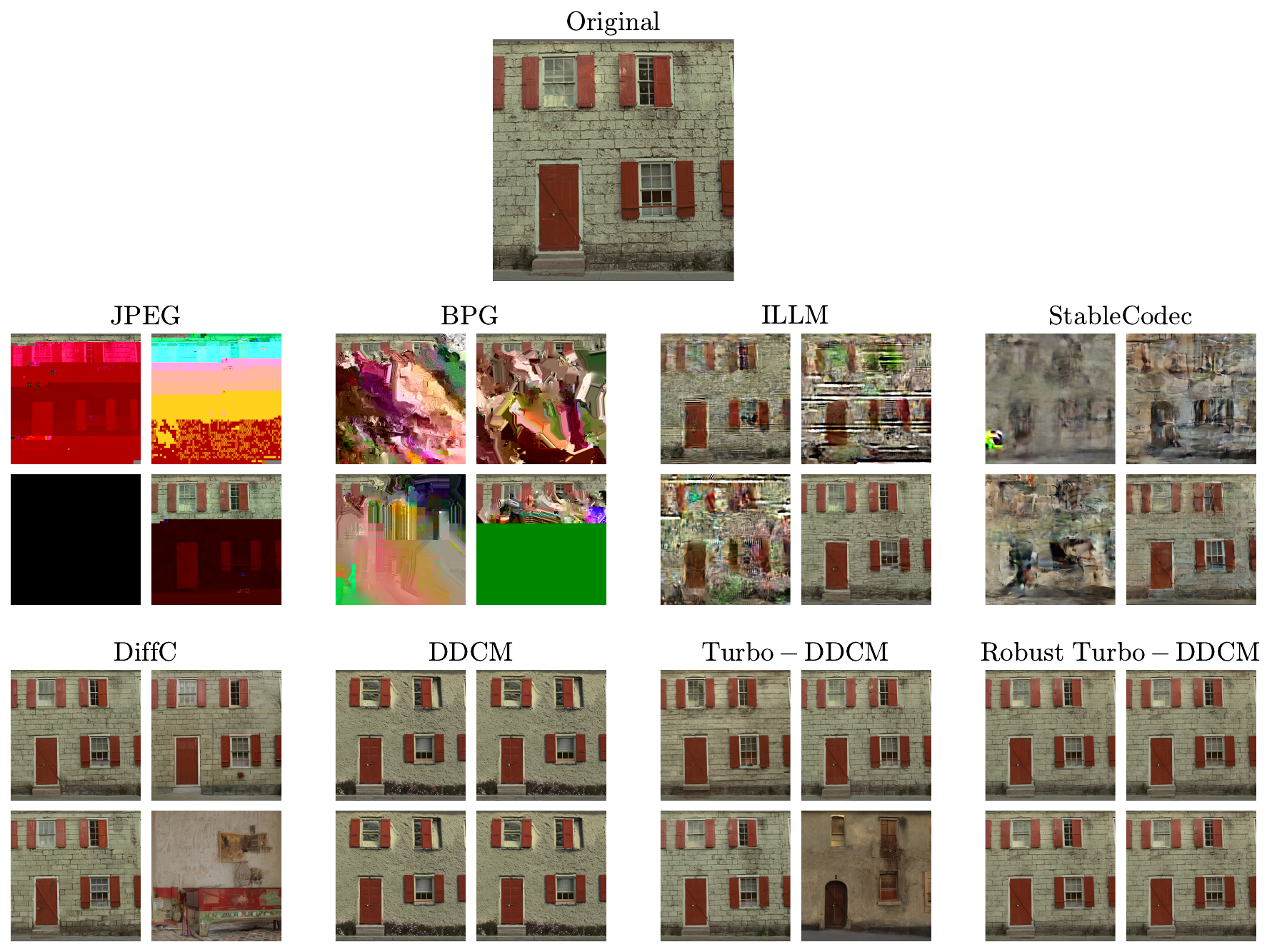}
    \includegraphics[width=0.8\textwidth]{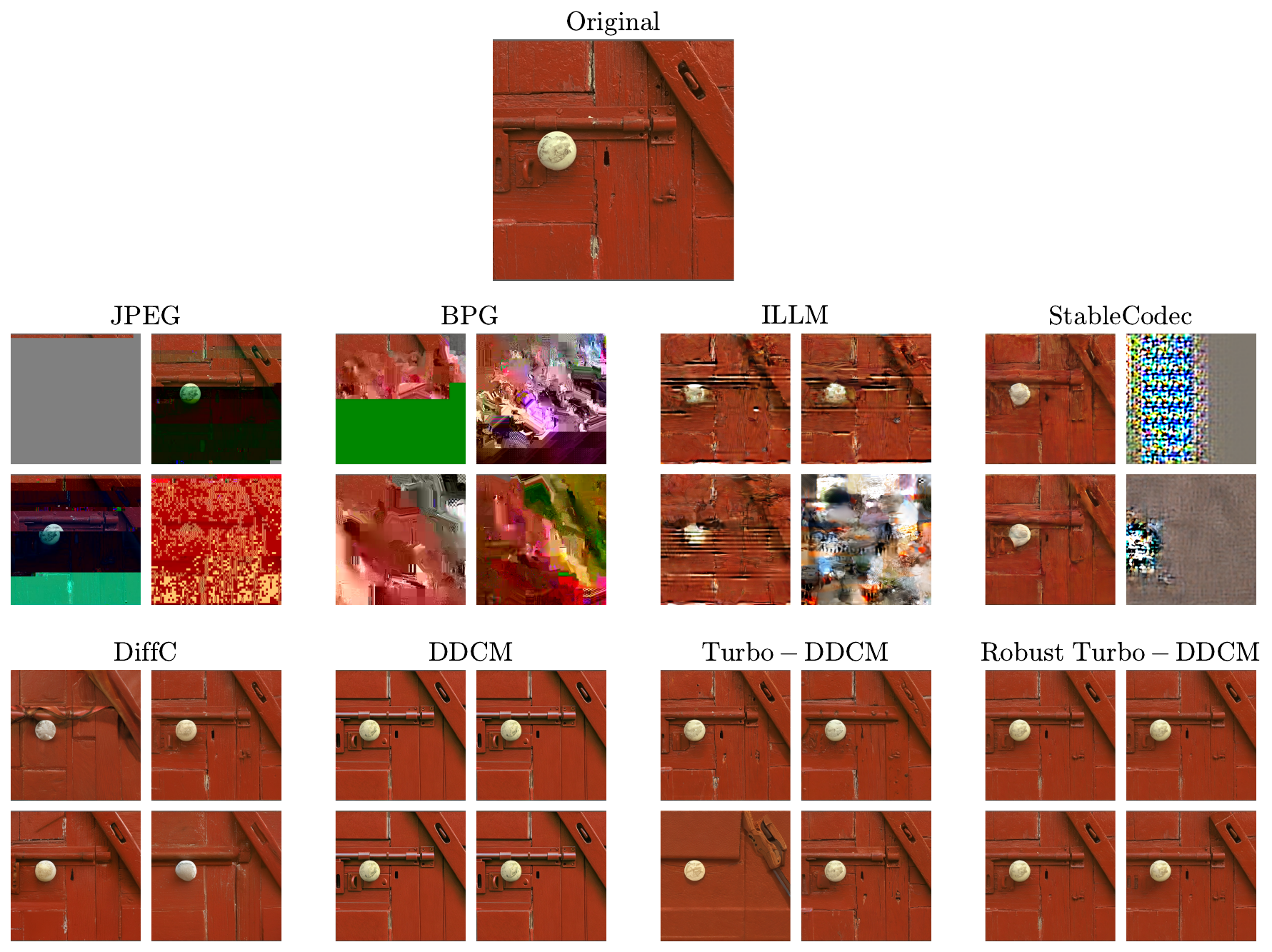}
    \caption{\textbf{Qualitative Results:} The presented reconstructions are based on images from the Kodak24 ($512 \times 512$) dataset under a BER of $10^{-4}$. While other methods produce reconstructions that aren't resemble the original image, RCC-based methods maintains high fidelity.}
    \label{fig:qaul_10_minus_4}
\end{figure*}

\begin{figure*}[t]
    \centering
    \includegraphics[width=0.8\textwidth]{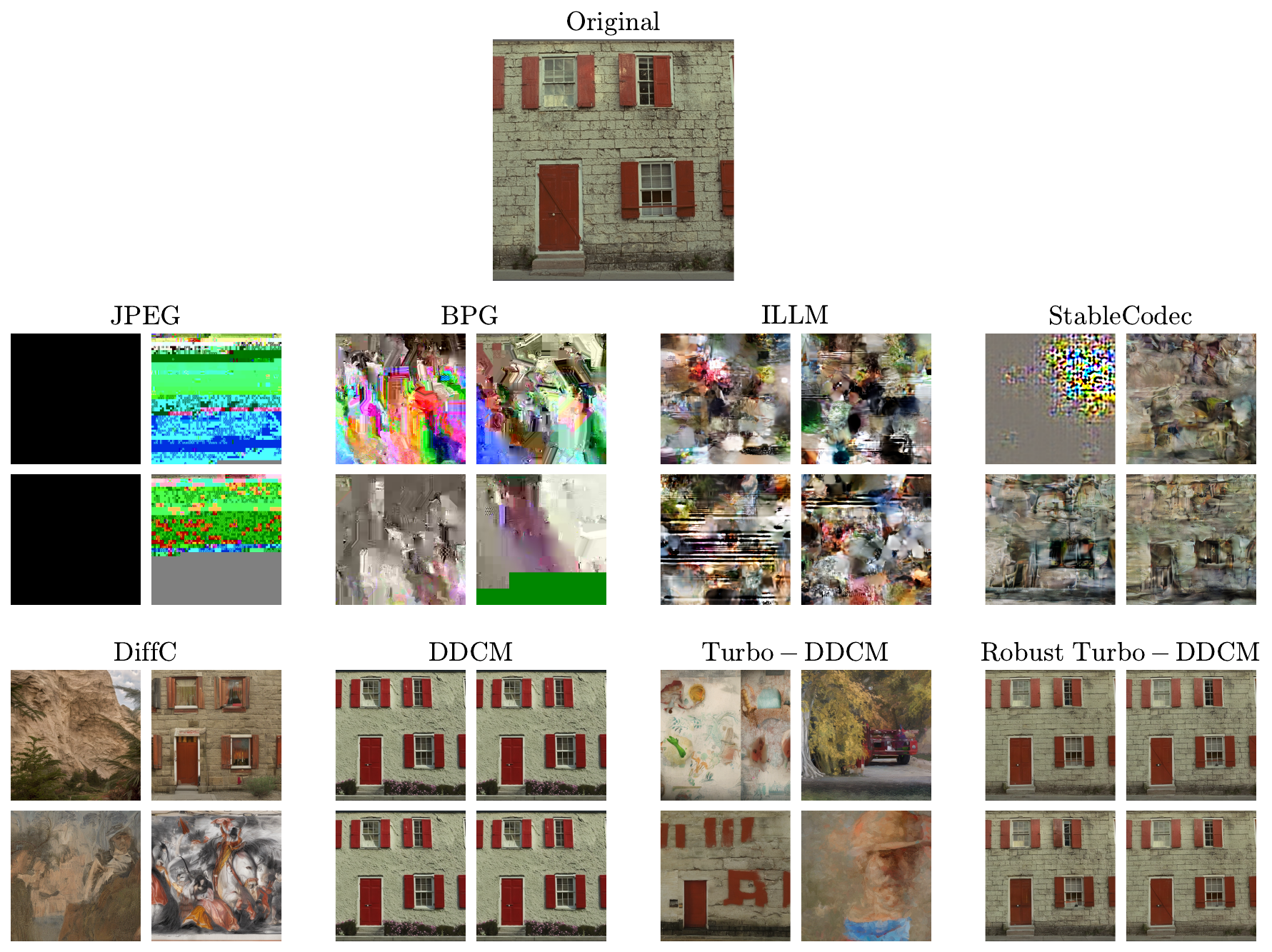}
    \includegraphics[width=0.8\textwidth]{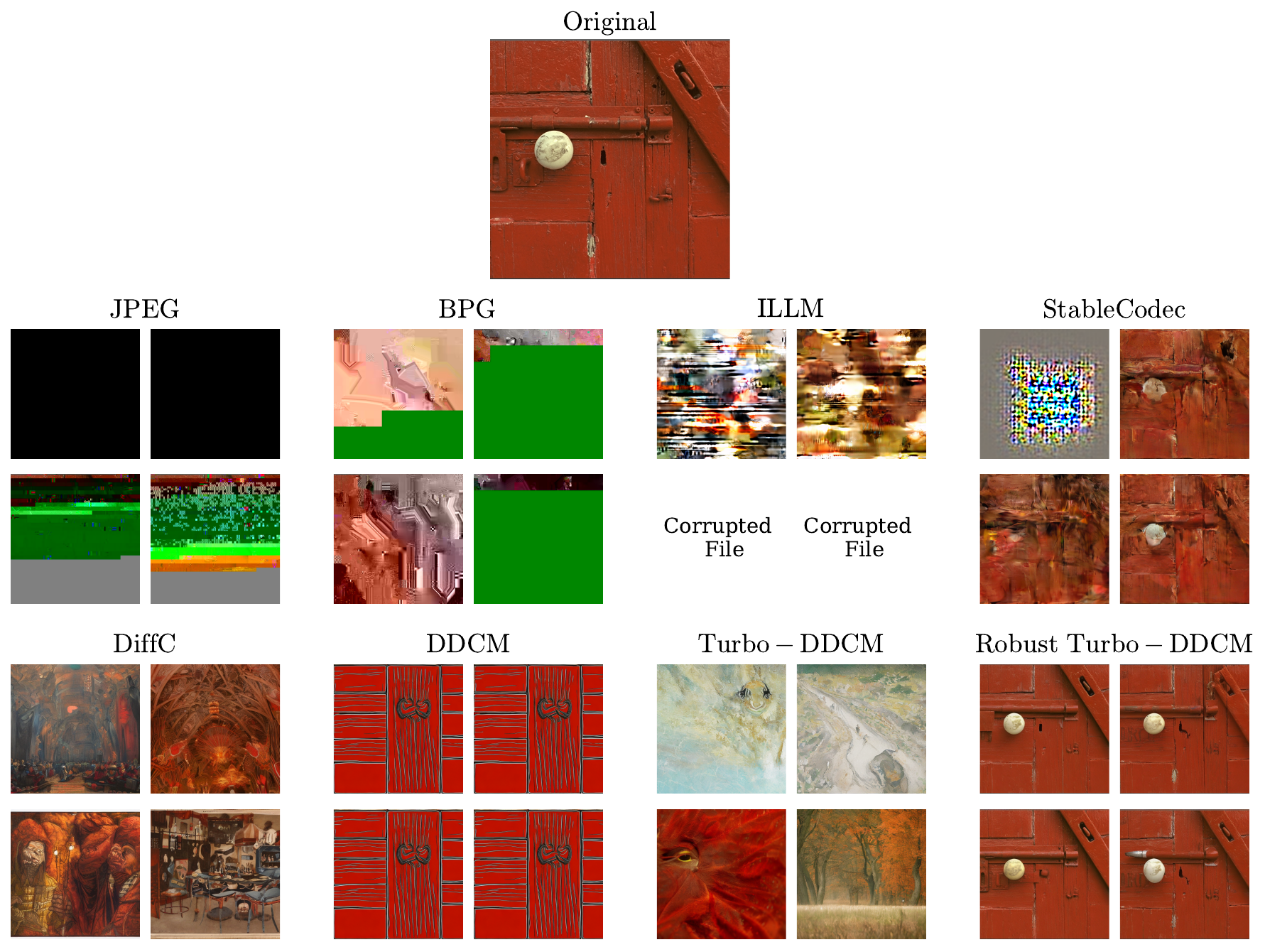}
    \caption{\textbf{Qualitative Results:} The presented reconstructions are based on images from the Kodak24 ($512 \times 512$) dataset under a BER of $10^{-3}$. While other methods produce reconstructions that no longer resemble the original image, our method maintains high visual fidelity.}
    \label{fig:qaul_10_minus_3}
\end{figure*}

% \raz{Problem setup is missing, you need to define the corruption model precisely: what BER means, whether flips are i.i.d. Bernoulli over the transmitted bitstream, how corrupted files are counted, and how undecodable outputs are treated in evaluation.}

% \gal{gal: i defined it earlier but im not sure if to seprate 3.1 with 4.1}

% \raz{It should be here, a subsection for experimental setup. Instea of expeirmental setting, use paragraphs in Experimental Setup for baselines, metrics, etc...}

\subsection{Experimental Setting}
\label{sec:exp-setting}

% We evaluate our analysis assumption and robust compression method on the Kodak24~\citep{franzen1999kodak} and DIV2K~\citep{agustsson2017ntire} datasets, using center-cropped images of size $512\times512$. We compare our approach with several diffusion-based RCC methods, including DDCM~\citep{ohayon2025compressed}, Turbo-DDCM~\citep{vaisman2025turboddcmfastflexiblezeroshot}, and DiffC~\citep{theis2022lossy}, implemented using the custom CUDA kernel of DiffC~\citep{vonderfecht2025lossy}. In addition, we compare against non-neural codecs such as JPEG and BPG~\citep{bpg}, the VAE-based method ILLM~\citep{muckley2023improving}, and the one-step diffusion training-based approach StableCodec~\citep{Zhang_2025_ICCV}. Detailed configurations are provided in~\todo{}. Distortion is evaluated using PSNR and LPIPS~\citep{zhang2018perceptual}, while perceptual quality is assessed using FID~\citep{bińkowski2018demystifying}, computed on $64\times64$ patches following~\citet{mentzer2020high}.

We perform our robustness analysis and evaluate our proposed method on the Kodak24~\citep{franzen1999kodak} and DIV2K~\citep{agustsson2017ntire} datasets, using center-cropped images of size $512\times512$. We compare against several diffusion RCC-based methods, including DDCM~\citep{ohayon2025compressed}, Turbo-DDCM~\citep{vaisman2025turboddcmfastflexiblezeroshot}, and DiffC~\citep{theis2022lossy}, implemented using the custom CUDA kernel of DiffC~\citep{vonderfecht2025lossy}. We also compare against non-neural codecs such as JPEG~\citep{wallace1991jpeg} and BPG~\citep{bpg}, the hybrid autoencoder--GAN method ILLM~\citep{muckley2023improving}, and the one-step diffusion-based approach StableCodec~\citep{Zhang_2025_ICCV}. Based on preliminary visual inspection of reconstruction quality, we evaluate JPEG at $1.0$ BPP, BPG at $0.5$ BPP, and the neural-based methods at approximately $0.1$ BPP. Full configurations, bitrate-selection details, and method-specific hyperparameters are provided in Table~\ref{app:configs}. Distortion is measured using PSNR and LPIPS~\citep{zhang2018perceptual}, while perceptual quality is evaluated with FID~\citep{bińkowski2018demystifying}, computed on $64\times64$ patches following~\citet{mentzer2020high}.

\subsection{Experimental Results}
\label{sec:exp-results}
As shown quantitatively in Fig.~\ref{fig:quant_main} and qualitatively in Figs.~\ref{fig:qaul_10_minus_4} and~\ref{fig:qaul_10_minus_3}, diffusion RCC-based methods consistently demonstrate superior robustness to bit flips across all datasets and metrics. While conventional codecs and learned compression methods rapidly degrade even at low BER, RCC-based approaches maintain stable performance over a wide noise range. In particular, our Robust Turbo-DDCM exhibits near-immunity to channel noise, with minimal degradation observed up to BER $10^{-3}$, where all other methods fail.

PSNR for non-RCC methods drops sharply already at BER $\sim 10^{-5}$--$10^{-4}$, whereas RCC-based methods degrade much more gradually. Our method consistently achieves the highest PSNR across almost all BER values. Similarly, FID increases rapidly for standard methods, reflecting severe distributional shifts, while RCC-based methods exhibit significantly slower growth. Our robust variant maintains the lowest FID for almost all BER values, indicating better preservation of global image structure and semantics under noise.

The gap is most pronounced in the ``\% Corrupted Files'' metric. Non-RCC methods undergo a sharp failure transition, reaching more than 80\% corrupted outputs around BER $10^{-2}$. In contrast, RCC-based methods largely avoid failures, with our approach maintaining zero corrupted files across the entire BER range. These results highlight the robustness of RCC-based compression and establish our method as a reliable solution for transmission over noisy channels.

We further analyze the rate--distortion--perception trade-off of RCC-based methods in Fig.~\ref{fig:rate_distortion_perception}. While Turbo-DDCM and DiffC achieve the best performance in terms of distortion and perceptual quality at a given bitrate, Robust Turbo-DDCM exhibits a modest degradation due to its more redundant encoding scheme. This reflects an inherent trade-off: increased robustness to bit flips comes at the cost of a slight reduction in compression efficiency.

\section{Discussion}
\label{sec:discussion}

Our results highlight a relatively underexplored aspect of neural image compression: robustness to bit-level corruption. Although such errors are typically mitigated using ECC, our experiments show that diffusion-based methods built on the RCC paradigm are more robust than classical and learned codecs across multiple noise levels. More broadly, these findings suggest that RCC-based compression may motivate moving beyond the standard pipeline of first compressing and then separately protecting the bitstream. Because the compressed representation itself is more resilient to corruption, it may be possible to use weaker ECC while still maintaining acceptable reconstructions when some bit errors remain.

In addition, we show that the encoding protocol itself plays an important role in robustness. In Turbo-DDCM, a single corrupted lexicographic index may change the entire selected atom set, making the representation sensitive to bit flips. Robust Turbo-DDCM mitigates this issue by encoding atom indices independently, which localizes the impact of bit errors. This modification has only a small impact on the rate--distortion--perception trade-off, while substantially improving robustness to bit-level corruption.

Our study has several limitations. First, we evaluate robustness using a Binary Symmetric Channel with independent bit flips, whereas real communication channels may exhibit burst or other structured errors. Second, some of the evaluated methods use entropy coding, which increases sensitivity to bit flips, whereas DDCM, Turbo-DDCM, and Robust Turbo-DDCM do not. This makes it difficult to fully disentangle the contribution of the underlying representation from that of the encoding scheme. Still, we observe robustness differences both among entropy-coded methods and among non-entropy-coded methods, suggesting that the advantage of Robust Turbo-DDCM is not solely due to the absence of entropy coding. These limitations could be addressed in future work.

\FloatBarrier

\bibliographystyle{ieeenat_fullname}
\bibliography{main}

\clearpage
\appendix
% ===== APPENDIX =====

\clearpage
\onecolumn

\section*{Appendix}

\section{Experimental Configurations and Additional Results}
\label{app:turbo-ddcm-rate-distortion-tradeoff}

In this section, we first summarize the configurations of all evaluated compression methods in Table~\ref{app:configs}. To determine the target bitrate for each method, we first performed a manual search for operating points that did not introduce clear visible artifacts in the reconstructed images. Based on this preliminary inspection, we used higher bitrates for the classical codecs, namely $1.0$ BPP for JPEG and $0.5$ BPP for BPG. For the remaining neural-based methods, we targeted a bitrate of approximately $0.1$ BPP, which provided a reasonable trade-off between compression and perceptual quality. In addition, Tables~\ref{tab:bitflip_results_1e4} and~\ref{tab:bitflip_results_1e3} provide the exact results from Fig.~\ref{fig:quant_main}.

\begin{center}
\footnotesize
\captionsetup{type=table}
\caption{Experimental configurations for the evaluated compression methods.}
\label{app:configs}
\begin{tabular}{L{0.10\textwidth} L{0.29\textwidth} L{0.18\textwidth} L{0.25\textwidth} L{0.08\textwidth}}
\toprule
Method & Representation & Hyperparameters & BPP setting & Entropy \\
\midrule

JPEG
& Quantized DCT coefficients from a fixed transform on each $8\times8$ block
& Quality parameter $q$
& Target BPP $=1.0$; per-image binary search on $q$
& Yes \\

BPG
& Block prediction information and quantized HEVC residual transform coefficients
& Quality parameter $q$
& Target BPP $=0.5$; per-image binary search on $q$
& Yes \\

ILLM
& Quantized learned encoder outputs
& Pretrained model \texttt{msillm\_quality\_2}
& Fixed by pretrained model ($0.07$ BPP)
& Yes \\

StableCodec
& Quantized learned latent features decoded via one-step diffusion
& \texttt{stablecodec\_ft2} checkpoint (highest available publicly)
& $\approx 0.035$ BPP; fixed by pretrained model
& Yes \\

DiffC
& RCC-based seeds defining Gaussian noise for each image patch and reverse-diffusion step
& Stable Diffusion 2.1; reconstruction timestep 20
& $\approx 0.1$ BPP
& Yes \\

DDCM
& RCC-based indices of selected Gaussian codebook atoms with coefficients
& Stable Diffusion 2.1; $T=1000$, $K=8192$, $M=2$, $C=3$
& Fixed image size; binary search over $M$ for $\approx 0.1$ BPP
& No \\

Turbo-DDCM
& RCC-based indices of selected Gaussian codebook atoms with coefficients; atoms jointly encoded as one lexicographic index over $\binom{K}{M}$ subsets
& Stable Diffusion 2.1; $T=30$, $K=16384$, $M=114$, $C=1$
& Fixed image size; binary search over $M$ for $\approx 0.1$ BPP
& No \\

Robust Turbo-DDCM
& RCC-based indices of selected Gaussian codebook atoms with coefficients; unlike Turbo-DDCM, the $M$ atoms are encoded separately
& Stable Diffusion 2.1; $T=30$, $K=16384$, $M=73$, $C=1$
& Fixed image size; binary search over $M$ for $\approx 0.1$ BPP
& No \\

\bottomrule
\end{tabular}
\end{center}

\footnotesize
\setlength{\LTcapwidth}{\textwidth}

\begin{longtable}{L{0.11\textwidth} L{0.19\textwidth} L{0.12\textwidth} L{0.12\textwidth} L{0.12\textwidth} L{0.15\textwidth}}
\caption{\textbf{Exact aggregated results for all methods at BER $10^{-4}$.} The table reports bit-flip robustness on Kodak24 and DIV2K using PSNR, LPIPS, FID, and the percentage of corrupted output files. Values are mean $\pm$ standard deviation, except corrupted files, which are reported as percentages. Bold indicates the best result for each metric and dataset.}
\label{tab:bitflip_results_1e4}\\
\toprule
Dataset & Method & PSNR $\uparrow$ & LPIPS $\downarrow$ & FID $\downarrow$ & \makecell{\% Corrupted\\Files $\downarrow$} \\
\midrule
\endfirsthead

\toprule
Dataset & Method & PSNR $\uparrow$ & LPIPS $\downarrow$ & FID $\downarrow$ & \makecell{\% Corrupted\\Files $\downarrow$} \\
\midrule
\endhead

\bottomrule
\endfoot

Kodak24 & JPEG & 9.15 $\pm$ 2.54 & 0.75 $\pm$ 0.17 & 157.98 $\pm$ 24.56 & 0.83 \\
 & BPG & 10.43 $\pm$ 2.00 & 0.75 $\pm$ 0.11 & 120.03 $\pm$ 4.26 & 2.08 \\
 & ILLM & 17.43 $\pm$ 5.45 & 0.42 $\pm$ 0.22 & 75.05 $\pm$ 5.25 & 7.08 \\
 & StableCodec & 19.12 $\pm$ 4.81 & 0.38 $\pm$ 0.28 & 57.30 $\pm$ 6.82 & 0.42 \\
 & DiffC & 20.92 $\pm$ 4.18 & 0.29 $\pm$ 0.18 & 33.95 $\pm$ 1.74 & 0.87 \\
 & DDCM & 20.52 $\pm$ 3.70 & 0.30 $\pm$ 0.18 & 40.31 $\pm$ 6.17 & \textbf{0.00} \\
 & Turbo-DDCM & 21.68 $\pm$ 4.18 & 0.25 $\pm$ 0.18 & 31.35 $\pm$ 2.53 & 0.42 \\
 & Robust Turbo-DDCM & \textbf{24.05 $\pm$ 2.90} & \textbf{0.15 $\pm$ 0.07} & \textbf{27.46 $\pm$ 0.19} & \textbf{0.00} \\
\midrule
DIV2K & JPEG & 8.52 $\pm$ 2.48 & 0.72 $\pm$ 0.17 & 173.06 $\pm$ 7.18 & 2.20 \\
 & BPG & 9.13 $\pm$ 1.84 & 0.74 $\pm$ 0.12 & 150.19 $\pm$ 2.27 & 1.50 \\
 & ILLM & 15.08 $\pm$ 5.11 & 0.45 $\pm$ 0.20 & 106.93 $\pm$ 4.87 & 9.70 \\
 & StableCodec & 17.01 $\pm$ 4.71 & 0.40 $\pm$ 0.28 & 61.96 $\pm$ 5.10 & 1.70 \\
 & DiffC & 18.62 $\pm$ 4.14 & 0.30 $\pm$ 0.18 & 37.27 $\pm$ 1.29 & 0.80 \\
 & DDCM & 19.04 $\pm$ 3.56 & 0.28 $\pm$ 0.17 & 43.12 $\pm$ 1.64 & \textbf{0.00} \\
 & Turbo-DDCM & 20.17 $\pm$ 4.12 & 0.24 $\pm$ 0.17 & 34.78 $\pm$ 1.59 & 0.20 \\
 & Robust Turbo-DDCM & \textbf{22.03 $\pm$ 3.07} & \textbf{0.17 $\pm$ 0.08} & \textbf{31.54 $\pm$ 0.19} & \textbf{0.00} \\
\end{longtable}

\begin{longtable}{L{0.11\textwidth} L{0.19\textwidth} L{0.12\textwidth} L{0.12\textwidth} L{0.12\textwidth} L{0.15\textwidth}}
\caption{\textbf{Exact aggregated results for all methods at BER $10^{-3}$.} The table reports bit-flip robustness on Kodak24 and DIV2K using PSNR, LPIPS, FID, and the percentage of corrupted output files. Values are mean $\pm$ standard deviation, except corrupted files, which are reported as percentages. Bold indicates the best result for each metric and dataset.}
\label{tab:bitflip_results_1e3}\\
\toprule
Dataset & Method & PSNR $\uparrow$ & LPIPS $\downarrow$ & FID $\downarrow$ & \makecell{\% Corrupted\\Files $\downarrow$} \\
\midrule
\endfirsthead

\toprule
Dataset & Method & PSNR $\uparrow$ & LPIPS $\downarrow$ & FID $\downarrow$ & \makecell{\% Corrupted\\Files $\downarrow$} \\
\midrule
\endhead

\bottomrule
\endfoot

Kodak24 & JPEG & 7.84 $\pm$ 2.14 & 0.98 $\pm$ 0.08 & 420.95 $\pm$ 38.97 & 20.83 \\
 & BPG & 9.33 $\pm$ 1.56 & 0.84 $\pm$ 0.08 & 140.70 $\pm$ 6.28 & 15.83 \\
 & ILLM & 10.12 $\pm$ 2.78 & 0.75 $\pm$ 0.10 & 169.38 $\pm$ 9.16 & 61.67 \\
 & StableCodec & 13.18 $\pm$ 3.62 & 0.74 $\pm$ 0.21 & 129.21 $\pm$ 15.79 & 10.83 \\
 & DiffC & 14.85 $\pm$ 3.00 & 0.62 $\pm$ 0.15 & 57.94 $\pm$ 3.53 & 8.26 \\
 & DDCM & 16.74 $\pm$ 2.69 & 0.52 $\pm$ 0.16 & 76.57 $\pm$ 7.09 & \textbf{0.00} \\
 & Turbo-DDCM & 14.28 $\pm$ 3.17 & 0.66 $\pm$ 0.17 & 70.22 $\pm$ 4.81 & 2.92 \\
 & Robust Turbo-DDCM & \textbf{22.57 $\pm$ 2.58} & \textbf{0.20 $\pm$ 0.08} & \textbf{30.15 $\pm$ 0.57} & \textbf{0.00} \\
\midrule
DIV2K & JPEG & 7.47 $\pm$ 2.64 & 0.97 $\pm$ 0.10 & 456.82 $\pm$ 35.16 & 22.60 \\
 & BPG & 8.72 $\pm$ 1.77 & 0.80 $\pm$ 0.10 & 164.36 $\pm$ 2.57 & 17.00 \\
 & ILLM & 9.21 $\pm$ 2.59 & 0.72 $\pm$ 0.13 & 185.79 $\pm$ 7.23 & 64.10 \\
 & StableCodec & 11.28 $\pm$ 3.17 & 0.75 $\pm$ 0.21 & 135.96 $\pm$ 4.88 & 13.00 \\
 & DiffC & 13.48 $\pm$ 2.83 & 0.58 $\pm$ 0.15 & 55.33 & 5.20 \\
 & DDCM & 15.58 $\pm$ 2.67 & 0.47 $\pm$ 0.16 & 65.78 $\pm$ 2.94 & 0.20 \\
 & Turbo-DDCM & 13.41 $\pm$ 3.13 & 0.62 $\pm$ 0.18 & 77.59 $\pm$ 3.36 & 2.70 \\
 & Robust Turbo-DDCM & \textbf{20.74 $\pm$ 2.69} & \textbf{0.21 $\pm$ 0.09} & \textbf{34.71 $\pm$ 0.40} & \textbf{0.00} \\
\end{longtable}

% if the rest of the paper after appendix should return to two columns:
% \clearpage
% \twocolumn


\end{document}